\documentclass[preprint]{elsarticle}

\usepackage{lineno,hyperref}
\modulolinenumbers[5]

\usepackage{CJKutf8}
\usepackage[T1]{fontenc}

\usepackage{booktabs}
\usepackage{multirow}
\usepackage{tabularx}

\graphicspath{ {./figures/} }

\usepackage{color, soul}
\sethlcolor{white}

\newcommand\blfootnote[1]{%
  \begingroup
  \renewcommand\thefootnote{}\footnote{#1}%
  \addtocounter{footnote}{-1}%
  \endgroup
}

\begin{document}

\begin{frontmatter}

\title{Can You Fool AI by Doing a 180? -- A Case Study on Authorship Analysis of Texts by Arata Osada\blfootnote{© 2022. Licensed under the \href{https://creativecommons.org/licenses/by-nc-nd/4.0/}{Creative Commons BY-NC-ND}.}}

\author[mymainaddress]{Jagna Nieuwazny\corref{mycorrespondingauthor}}
\cortext[mycorrespondingauthor]{Corresponding author: jagna@e.koeki-u.ac.jp}

\author[mymainaddress]{Karol Nowakowski} 

\author[mysecondaryaddress]{Michal Ptaszynski}

\author[mysecondaryaddress]{Fumito Masui}

\address[mymainaddress]{Tohoku University of Community Service and Science, 3-5-1 Iimoriyama, Sakata, 998-0875, Yamagata, Japan.}

\address[mysecondaryaddress]{Kitami Institute of Technology, Koencho 165, Kitami, 090-8507, Hokkaido, Japan.}

\address{\textup{\texttt{jagna@e.koeki-u.ac.jp}, \texttt{karol@koeki-u.ac.jp}, \texttt{michal@mail.kitami-it.ac.jp}, \texttt{f-masui@mail.kitami-it.ac.jp}}}

\begin{abstract}

This paper is our attempt at answering a twofold question covering the areas of ethics and authorship analysis solutions.
Firstly, since the methods used for performing authorship analysis imply that an author can be recognized by the content he or she creates, we were interested in finding out whether it would be possible for an author identification system to correctly attribute works to authors if in the course of years they have undergone a major psychological transition. 
Secondly -- and from the point of view of the evolution of an author's ethical values -- we checked what it would mean if the authorship attribution system encounters difficulties in detecting single authorship. 
We set out to answer those questions through performing a binary authorship analysis task using a text classifier based on a pre-trained transformer model and a baseline method relying on conventional similarity metrics. 
For the test set, we chose several works of Arata Osada, a Japanese educator and specialist in the history of education, with half of them being books written before the Second World War and another half in the 1950s, in between which the author underwent a transformation in terms of political opinions.
As a result, we were able to confirm that in the case of texts authored by Arata Osada in a time span of more than 10 years, while the classification accuracy drops by a large margin and is substantially lower than for texts by other non-fiction writers, confidence scores of the predictions remain at a similar level as in the case of a shorter time span, indicating that the classifier was in many instances tricked into deciding that texts written by Arata Osada over a time span of multiple years were actually written by two different people, which in turn leads us to believe that such a change can affect authorship analysis, and that historical events have great impact on a person’s ethical outlook as expressed in their writings.
\vspace{1em}

\end{abstract}

\begin{keyword}

\texttt{authorship analysis, single authorship identification, authorship verification, similarity detection, binary text classification, transformers, personal ethics}

\end{keyword}

\end{frontmatter}


\section{Introduction}

Authorship analysis is the process of determining authorship of a document based on its characteristics. 
The problem itself has a long history, dated back to the end of the 19th century when Mendenhall \cite{Mendenhall237} examined for the first time the word length in the works of Bacon, Shakespeare and Marlowe in order to detect quantitative stylistic differences.

In a computational context, it is an emerging area of research associated with applications in literary research, cyber-security, forensics, and social media analysis. Other applications include:

\begin{itemize}
\item \textbf{Forensic linguistics and online bullying detection} (see \cite{Chaski2013BestPA,frommholz2017textual}) -- to identify characteristics of the author of anonymous, pseudonymous or forged text, based on the author’s use of the language (blackmailing letters, confessions, testaments, suicide letters).
\item \textbf{Bot detection} -- in the context of marketing, social bots can artificially inflate the popularity of a product by posting positive reviews. Especially Twitter bots can be considered a threat given their commercial, political and ideological influence, such as in the 2016 United States Presidential Election \cite{Bovet}, during which they polarised political conversations, and spread fake news.
\item \textbf{Marketing} -- to identify the demographics of people that like or dislike their products based on the analysis of blogs, online product reviews and social media content (see \cite{Juola2015IndustrialUF}).

\end{itemize}

Much research has been focused on determining suitable features for modeling writing patterns from authors. Reported results indicate that content-based, as well as style-based features continue to be the most relevant and discriminant features for solving this task (see \cite{Sari2018TopicOS}). 

The remainder of this paper is organized as follows.
In the following section we introduce our research questions.
In Section~\ref{sec:background}, we present the background of our study, including previous research on authorship analysis.
We also discuss the influence of historical events on changes in ethical values and the cultural importance of the \textit{shinpoteki bunkanin},
\hl{Japanese intellectuals active after the Second World War in Japan, instrumental in disseminating democratic ideas and pacifism in Japan.}
We then introduce the person of Arata Osada, our object of analysis.
In Section~\ref{sec:materials}, we introduce the resources applied to perform this study.
Section~\ref{sec:system} describes the classification model we applied to our task, as well as the procedure of data generation.
Finally, we present the results of our experiments and conclude the paper with a discussion on the results and some ideas for future work.

\section{Research questions}
\label{sec:question}

Recently, Nieuwazny et al. \cite{nieuwazny} investigated whether it is possible to automatically determine if changes in ethical education influence core moral values in humans throughout the century, on the example of Japan. As a result, they found out that, despite the changes in stereotypical view on Japan’s moral sentiments, as well as the changes in the model of education, especially due to historical events (pre- and post-World War II), past and contemporary Japanese shared a similar moral evaluation of certain basic moral concepts. This was an interesting discovery, since it showed that Japan, known for its imperialist and militarist ethical profile before the war, which greatly influenced people's attitudes towards death (e.g., suicide airplane pilots \textit{kamikaze}), in fact had a set of core moral values, which included the importance of preservation of life, which it shared throughout the history despite its official profile as a country.

In this paper we build upon the above findings. In their work, Nieuwazny et al. \cite{nieuwazny} focused on factors influencing the ethical values at a macro-scale of nation-wide population, such as general ethical education policies.
In this study, on the other hand, we focused on analyzing the factors influencing not a global population, but rather an individual’s ethical values.

In particular, we wanted to know if -- given that education policies can only influence someone’s ethical outlook to a limited extent -- it is possible to specify other external stimuli, such as major historical events (e.g., the World War II), that would change the ethical profile of particular individuals.

The specific research question we set up to answer in this research was twofold. Firstly, since the methods used for performing authorship analysis imply that an author can be recognized by the written content he or she creates (Internet messages, articles, books, etc.), how accurate would they prove in the case of an author who diametrically changed their opinions due to an external stimulus, such as the impact of war?

From the point of view of authorship analysis, we wanted to know if it would be possible for an authorship analysis solution to correctly attribute works from before and after the war to the same author and with what accuracy.
Another thing we checked was if such a shift in opinions could be precisely quantified, and on what level it could be perceived.
Moreover, we checked how such a difference would compare to the situation where two completely different authors are taken into account.

From the point of view of the impact on a person’s ethical outlook as expressed in their writings, if the content of an author’s work changed in between before and after the war, and if the authorship analysis system encounters difficulties in detecting single authorship, it would mean that historical events can have a great impact on a person’s ethical outlook. 

\section{Research Background}
\label{sec:background}

\subsection{Previous research}

Authorship analysis is the task of examining the characteristics of a document in order to draw conclusions about its authorship \cite{zheng}.

A 1964 study by Mosteller and Wallace \cite{mosteller} on the authorship of ``The Federalist Papers'' (a series of 146 political essays written by John Jay, Alexander Hamilton, and James Madison, whose authorship was controversial), where Bayesian statistical analysis of word frequencies was used, initiated non-traditional authorship analysis studies, no longer relying on a human specialist to determine authorship.

Until the ascension of the Internet and social media, research in authorship analysis was dominated by a computer-assisted, but not computer-based approach relying on the process of identification of quantifiable language features, such as word length, phrase length, sentence length, vocabulary frequency, distribution of words of different lengths, etc. \cite{article},  known as \textit{stylometry}.
The limitations of this approach included, among others, a small number of candidate authors being taken into consideration and the lack of suitable benchmark data \cite{stamatatos}.

The propagation of the Internet and social media and the virtually immeasurable amount of electronic texts available through it proved to be a great stimulus for the evolution of Natural Language Processing, Information Retrieval and Machine Learning, and as a consequence authorship analysis solutions as well. At the same time, the amount of text information to process and categorize indicated the potential of authorship analysis in several applications \cite{madigan} and the need for a reliable, computational method to perform it. 
In the last decade, areas of research in connection to authorship analysis include efforts to develop practical applications dealing with real world texts rather than solving literary questions.

In the typical authorship analysis problem -- known as closed-set authorship identification or authorship attribution -- a text of unknown authorship is assigned to one candidate author, given a set of candidate authors for whom text samples of undisputed authorship are available.
In open-set attribution, on the other hand, the true author is not necessarily included in the set of candidate authors.
A special case of open-set attribution is authorship verification, where, given one or more documents by a single author and another, anonymous document, the task is to determine if that document was also written by the same author \cite{Potha2014,koppel2014}.

Existing approaches to authorship analysis can be divided in two main groups: \hl{those based on hand-crafted similarity metrics} and machine learning-based methods \cite{koppel2014} (\hl{that being said, machine learning approaches often also employ distance metrics, either as input features} \cite{Hrlimann2015GLADGL} \hl{or in the objective function} \cite{Boenninghoff2019ExplainableAV}).

Recently, a number of studies have been carried out on cross-domain authorship identification \cite{Kestemont2018OverviewOT} (where the texts of known and unknown authorship belong to different domains) and style change detection (where single-author and multi-author texts are to be distinguished), featuring several methods involving the use of n-grams \cite{sapkota-etal-2015-character} and deep learning \cite{bagnall,Qian2017DeepLB,inproceedings}.
Nirkhi et al. \cite{Nirkhi} investigated the effect of increasing the number of authors on an SVM-based authorship identification system's performance.

Azarbonyad et al.~\cite{azarbonyad2015_time_aware} analyzed the changes in word usage by authors of tweets and emails and proposed a similarity-based, time-aware authorship attribution approach.
We are not aware of any previous research investigating the effect of temporal changes in the context of machine learning-based authorship analysis.

Outside of the domain of automatic authorship attribution systems, Rexha et al. \cite{Rexha2018} conducted a study to determine if human evaluators can identify authorship among texts with high content similarity, and what features influence their decisions.

Concerning research involving the use of pre-trained language models (such as BERT) in authorship analysis, Barlas et al. \cite{Barlas} extended the successful authorship verification approach of Bagnall~\cite{bagnall}, based on a multi-headed classifier, by combining it with four different types of pre-trained language models.
Most recently, Shimizu \cite{shimizu} reported the results of an authorship identification experiment for Japanese, using data obtained from the Aozora Bunko (an open-source repository of Japanese literature, also used in this research) and BERT.

\subsection{Influence of historical events on change in ethical values}

As has been discussed in political studies, such as the one by Maja Zehfuss, \cite{Zehfuss}, ethics, especially in the form of specifically manipulated code of ethics taught to people, often becomes a strong motivation for war. In such cases, ethical considerations do not act as a constraint, on the contrary making a commitment to ethics enables war and enhances its violence. However, war is also an  agent of major changes -- wars may change individuals' value systems and influence future choices, as, for example, when individual war experiences shift their evaluation of costs and benefit, war-weariness (and by consequence a sharp turn towards pacifism) is one such effect. Finally, wars may entail structural changes for actors, unchosen shifts in the context or environment within which they act\footnote{\url{http://www.grandstrategy.net/Articles-pdf/evaluating_war.pdf}}. 
We chose the Second World War as a cut-off point in this research due to its major impact on the Japanese society as a whole and the Japanese intelligentsia\footnote{artists, teachers, academics, writers, etc.} specifically, who, as mentioned by John Dower \cite{dower1999embracing}, ``performed a virtuoso turnabout, since only a precious few opposed the war and virtually any author wanting to be published in the 1930s had to be a military enthusiast and a supporter of the idea of the expansion of Japan in Asia.''

\subsection{Cultural importance of \textit{shinpoteki bunkanin} in Japan}

\textit{Shinpoteki bunkanin} (\begin{CJK}{UTF8}{ipxm}進歩的文化人\end{CJK}, ``progressive men of letters'') is a term used to describe Japanese intellectuals active after the Second World War, who were instrumental in disseminating democratic ideas and pacifism in Japan.
However, the same people often supported Japanese militarism in the 1930s and quickly changed sides upon Japan’s defeat, often trying to conceal or destroy their pre-war publications, of which many reflect their peculiar shift of political opinions.

For this reason, to analyze how a major life event such as war influences one's ethical values we chose one of the representatives of \textit{shinpoteki bunkanin}, namely Arata Osada, \hl{whose} literary works made a clear and major turn after the war, from a militarist to an extreme pacifist attitude. Moreover, since many literary works of pre- and post war writers are freely available as a language resource, it is possible to quantify how exactly such attitudes changed after the war. To perform such quantification, in this research we apply a neural language model-based method for single authorship identification.

\subsection{The person of Arata Osada: object of analysis}

Arata Osada (\begin{CJK}{UTF8}{ipxm}長田新\end{CJK}, 1887-1961) was a Japanese educator, honorary professor of Hiroshima University and specialist in the history of education. Before and during the Second World War he became known for his vocal views on patriotic education based on the German model \hl{(}\textit{Nationalpädagogik} \hl{known in Japanese as} \textit{Kokka kyōikugaku}, \begin{CJK}{UTF8}{ipxm}国家教育学\end{CJK}, or ``National education'', 1944) as well as newspapers articles with a militaristic undertone\footnote{An example excerpt: ``War is the motivation for the advance of humanity. Japanese want to be reborn 7 times to serve their country and this is so out of their great love for the motherland. The Japanese army is powerful because each of them wants to be like a living shield to the Emperor and sacrifice their lives. Perfecting the army's education is like perfecting one's life [...]. Soldiers are determined to sacrifice their lives without regret for one ``absolute person'' [i.e., the emperor]'' -- excerpts from an article in the journal \textit{Seishonen Shidō} (\begin{CJK}{UTF8}{ipxm}青少年指導\end{CJK}, ``Instructing of youth and children'') from February 1944.}. On August 6, 1945, the atomic bomb was dropped on Hiroshima, in which attack Osada was seriously injured. In 1947, he became the first chairman of the Japanese Educational Research Association and was a professor at the University of Hiroshima until his retirement in 1953. He became one of the key players in post-war Japanese education reconstruction, including forming the Japanese Children's Association and serving as its first president. Based on his experience of the atomic bomb, Osada actively participated in peace movements opposing nuclear arms, and collected the notes of boys and girls who experienced the dropping of the atomic bomb and published it as \textit{Genbaku no ko -- Hiroshima no shōnen shōjo no uttae} (\begin{CJK}{UTF8}{ipxm}原爆の子〜広島の少年少女のうったえ\end{CJK}, ``Children of the Atomic Bomb -- The Plight of Boys and Girls of Hiroshima''). He made a public appeal to abolish the imperial system in 1956 in the magazine \textit{Kyōiku} (\begin{CJK}{UTF8}{ipxm}教育\end{CJK}, \textit{Education}, in September, 1956), saying ``[...] abolish the imperial system, the defeat in the war was a win in this sense -- it will become a path to democracy''.
His other post-war publications include \textit{Heiwa wo motomete} (\begin{CJK}{UTF8}{ipxm}平和を求めて\end{CJK}, ``In search for Peace'', 1962) and \textit{Shakaishugi no bunka to kyōiku -- watakushi no mita Soren to Chūgoku} (\begin{CJK}{UTF8}{ipxm}社会主義の文化と教育 わたくしのみたソ連と中国\end{CJK}, ``Socialist culture and education -- the Soviet Union and China as I saw it'', 1956). 

We chose his works as a basis to create our sample in this research due to the fact that most of his writing is centered around the subject of education of the youth (thus the topical content or spectrum remain unchanged in pre- and post-war publications), but the consequences of the Second World War seem to have brought on a major shift in his belief system: from loyal imperialist and militarist to an extreme pacifist. 

\section{Materials}
\label{sec:materials}

\paragraph{Aozora Bunko} 

\textit{Aozora Bunko}\footnote{\url{https://www.aozora.gr.jp/}} (\begin{CJK}{UTF8}{ipxm}青空文庫\end{CJK}, literally the ``Blue Sky Library'', also known as the ``Open Air Library'') is a Japanese digital library that encompasses thousands of works of Japanese-language fiction and non-fiction, including out-of-copyright books or works that the authors wish to make freely available.
Aozora Bunko was created on the Internet in 1997 to provide broadly available, free access to Japanese literary works whose copyrights had expired. Most of the texts provided are Japanese literature, and some translations from English literature. The resources are searchable by category, author, or title. The files can be downloaded in PDF format or simply viewed in HTML format.
In 2013, the Future of Books Fund (\textit{Hon no mirai kikin}, \begin{CJK}{UTF8}{ipxm}本の未来基金\end{CJK}) was established independently to assist funding and operations for Aozora Bunko.
As of January 5, 2019, Aozora Bunko includes more than 15,100 works, a majority of which are novels.

\paragraph{Books by Arata Osada}

As the main test data, we used four books authored or co-authored by Arata Osada, two among which were written before the Second World War: \textit{Kyōiku shisōshi} (\begin{CJK}{UTF8}{ipxm}教育思想史\end{CJK} \cite{kyoikushisoshi}, ``The history of educational thought''), published in 1931, and \textit{Shinkyōiku no kōsō -- Amerika no bunka-kyōiku wo hihan shite} (\begin{CJK}{UTF8}{ipxm}新教育の構想 アメリカの文化・教 育を批判して\end{CJK} \cite{shinkyoiku}, ``The making of a new education -- Critical thoughts on American culture and education''), written before and during the war and published in 1949; and two others written after the war: \textit{Nihon no unmei to kyōiku} (\begin{CJK}{UTF8}{ipxm}日本の運命と教育\end{CJK} \cite{unmei}, ``Japan’s destiny and education''), published in 1953, and \textit{Kyōiku kihonhō} (\begin{CJK}{UTF8}{ipxm}教育基本法\end{CJK} \cite{kihonho}, ``Basic Law of Eduaction''), published in 1957.
\hl{All four books share a common topic, namely, education.}

As a first step in preparing our data sample, we performed an OCR of the four above-mentioned books and in the case of books co-authored by Osada, selected only the fragments that he had authored. 

\section{Same authorship detection system}
\label{sec:system}

We set out to answer our research questions through performing an authorship analysis experiment, using a binary text classification model based on a pre-trained language model (BERT), trained to predict if two fragments of text presented to it were produced by the same author or two different people.

Recent years have witnessed major improvements in a number of Natural Language Processing benchmarks, owing to the advent of deep pre-trained language models (examples include text classification \cite{NIPS2019_8812_XLNet}, text summarization \cite{zhang2019pegasus,yan2020prophetnet}, question answering \cite{Peters:2018,devlin2018bert} and machine translation \cite{lample2019cross}).
One of them is BERT (Bidirectional Encoder Representations from Transformers), proposed by Devlin et al. \cite{devlin2018bert}.
One of the two tasks used in pre-training of a BERT model is next sentence prediction, where the model learns to predict whether two sentences are likely to occur next to each other in a corpus, or are unrelated.
This makes BERT a natural choice for our problem, which is similar.

We employed the Japanese version of BERT, released by the T\={o}hoku University's Inui Laboratory\footnote{\url{https://github.com/cl-tohoku/bert-japanese}} (specifically, the variant using both MeCab and WordPiece tokenization, without whole word masking).
The model was pre-trained on Japanese Wikipedia articles.

\subsection{Data}

The training set for our classifier consists of randomly picked samples from the Aozora Library, where each sample is made of two paragraphs, separated by a special token (\textit{[SEP]}).
50\% of the data set is comprised by positive samples (i.e., those where both paragraphs have the same author), with the label set to ``1''.
The remaining half are negative samples, with the ``0'' label.
Positive samples are further divided in three sub-categories of equal size:

\begin{itemize}
\item two paragraphs from the same document;
\item two paragraphs from different documents by the same author, where the time lapse between the publication of the first and the second document is less than or equal to 10 years;
\item two paragraphs from different documents by the same author, where the time lapse between the publication of the first and the second document is more than 10 years.
\end{itemize}

Negative samples are subdivided in two equally sized parts:

\begin{itemize}
\item paragraphs from two documents by different authors, where the time lapse between the publication of the first and the second document is less than or equal to 10 years;
\item paragraphs from two documents by different authors, where the time lapse between the publication of the first and the second document is more than 10 years.
\end{itemize}

We chose 10 years of time lapse as the cutting point in this experiment due to it being approximately the time between writing the last book in the pre-war sample and the first post-war book by Arata Osada\footnote{While it was published in 1949, \textit{Shinkyōiku no kōsō -- Amerika no bunka-kyōiku wo hihan shite} represents Osada's pre-war, nationalist beliefs. For this reason, in our experiment we treated it as a book from 1940, which is the estimated time of its writing.
For all other books in our data, we took into account the year of first publication.}.

Due to the maximum sequence length of 512 sub-word tokens, imposed by the pre-trained BERT model used in the experiment, longer paragraphs were truncated to keep the combined token count of both paragraphs in each sample within the limit.
In order to avoid significant differences in length between two paragraphs constituting a data point -- which might affect the system's performance -- we filtered out paragraphs with the number of tokens smaller than 200.
This left us with a total of 66,447 paragraphs in 6,111 documents by 412 authors\footnote{These statistics do not include three authors (namely, Yukichi Fukuzawa, Ky\={o}ka Izumi and Asajir\={o} Oka), whose works were excluded from the Aozora data set and used to build separate, single-author test sets for direct comparison with Arata Osada.}.
Authors were then split in three groups: (i) authors with only a single document, (ii) authors with multiple documents published within a period of 10 years and (iii) authors of multiple documents published over a period longer than 10 years.
After that, each group was randomly divided between the training set, development set, and test set, proportionally to the size of each data set.
Finally, we generated a specified number of data points for each data set and category of samples, by randomly sampling pairs of authors, their documents and paragraphs they comprise.

Test set composed of the four books by Arata Osada was compiled according to the same rules, the only difference being the fact that at least one of the paragraphs in each data point was sampled from his works (in the case of negative samples, the other paragraph was picked from the same pool of documents as those used in the Aozora Bunko test set).
Furthermore, we applied the same guidelines to create separate test sets for three other individual authors (excluded from the Aozora Bunko data set), with the aim of using them for direct comparison with Osada. These were: Yukichi Fukuzawa (\begin{CJK}{UTF8}{ipxm}福沢諭吉\end{CJK}, 1835-1901), an educator, entrepreneur, political scientist and translator of foreign literature\footnote{We included \hl{six of his works, all of them related to the topic of education}: \textit{Gakumon no susume} (\begin{CJK}{UTF8}{ipxm}学問のすすめ\end{CJK}, ``On learning'', 1872), \textit{Shōgaku kyōiku no koto} (\begin{CJK}{UTF8}{ipxm}小学教育の事\end{CJK}, ``On the subject of elementary education'', 1879), \textit{Kyōiku no mokuteki} (\begin{CJK}{UTF8}{ipxm}教育の目的\end{CJK}, ``The objective of education'', 1879), \textit{Dokurinri kyōkasho} (\begin{CJK}{UTF8}{ipxm}読倫理教科書\end{CJK}, ``Readings in ethics'', 1890), \textit{Onna daigaku hyōron} (\begin{CJK}{UTF8}{ipxm}女大学評論\end{CJK}, ``New essential learning for women'', 1899) and \textit{Shin onna daigaku} (\begin{CJK}{UTF8}{ipxm}新女大学\end{CJK}, ``New women university'', 1899).
}, Kyōka Izumi (\begin{CJK}{UTF8}{ipxm}泉鏡花\end{CJK}, 1873-1939), a novelist writing mostly about societal matters\footnote{We included: \textit{Katsu ningyō} (\begin{CJK}{UTF8}{ipxm}活人形\end{CJK}, ``Living doll'', 1893), \textit{Giketsu kyōketsu} (\begin{CJK}{UTF8}{ipxm}義血侠血\end{CJK}, ``Giketsukyōketsu'', 1894), \textit{Yakōjunsa} (\begin{CJK}{UTF8}{ipxm}夜行巡査\end{CJK}, ``Night patrol'', 1895), \textit{Bakeichō} (\begin{CJK}{UTF8}{ipxm}化銀杏\end{CJK}, ``The haunted gingko tree'', 1896), \textit{Ryūtandan} (\begin{CJK}{UTF8}{ipxm}竜潭譚\end{CJK}, ``A Deep Water's Dragon tale'', 1896), 
\textit{Kaijō hatsuden} (\begin{CJK}{UTF8}{ipxm}海城発電\end{CJK}, ``Kaijohatsuden'', 1896), 
\textit{Yushima mōde} (\begin{CJK}{UTF8}{ipxm}湯島詣\end{CJK}, ``Pilgrimage to Yushima'', 1899), 
\textit{Sanmai tsuzuki} (\begin{CJK}{UTF8}{ipxm}三枚続\end{CJK}, ``Sanmaitsuzuki'', 1900),
\textit{Kōgyoku} (\begin{CJK}{UTF8}{ipxm}紅玉\end{CJK}, ``Jonathan apple'', 1913), \textit{Nihonbashi} (\begin{CJK}{UTF8}{ipxm}日本橋\end{CJK}, ``Nihonbashi bridge'', 1914), \textit{Hinagatari} (\begin{CJK}{UTF8}{ipxm}雛がたり\end{CJK}, ``Hinagatari'', 1917),  \textit{Shippō no hashira} (\begin{CJK}{UTF8}{ipxm}七宝の柱\end{CJK}, ``Pillar of seven treasures'', 1917), 
\textit{Ōsaka made} (\begin{CJK}{UTF8}{ipxm}大阪まで\end{CJK}, ``To Osaka'', 1918) and \textit{Hakushaku no kanzashi} (\begin{CJK}{UTF8}{ipxm}伯爵の釵\end{CJK}, ``The earl's hair ornament'', 1920).} and Asajir\={o} Oka (\begin{CJK}{UTF8}{ipxm}丘浅次郎\end{CJK}, 1868-1944), a researcher writing about natural history\footnote{We used: \textit{Dōbutsu sekai ni okeru zen to aku} (\begin{CJK}{UTF8}{ipxm}動物界における善と悪\end{CJK}, ``Good and evil in the animal world'', 1902), \textit{Jinrui no kodaikyō} (\begin{CJK}{UTF8}{ipxm}人類の誇大狂\end{CJK}, ``The megalomania of the human kind'', 1904), \textit{Shizenkai no kyogi} (\begin{CJK}{UTF8}{ipxm}自然界の虚偽\end{CJK}, ``The falsehood of the natural world'', 1907), \textit{Seibutsugakuteki no mikata} (\begin{CJK}{UTF8}{ipxm}生物学的の見方\end{CJK}, ``From a biologist's point of view'', 1910), \textit{Warera no tetsugaku} (\begin{CJK}{UTF8}{ipxm}我らの哲学\end{CJK}, ``Our philosophy'', 1921) and \textit{Ningenseikatsu no mujun} (\begin{CJK}{UTF8}{ipxm}人間生活の矛盾\end{CJK}, ``The paradox of human life'', 1926).}.
\hl{The intention behind including works of Yukichi Fukuzawa and Asajir\={o} Oka was to verify how the classification results for Osada's works compare with those obtained for (i) an author who -- like Osada -- wrote multiple books on the topic of education (Yukichi Fukuzawa), and (ii) an author who also consistently wrote on a given topic, other than education (Asajir\={o} Oka and his books on natural history).
Furthermore, we included a novelist (Ky\={o}ka Izumi) in order to test the hypothesis that in the case of a fiction writer, covering a broad range of topics in between their works, the performance of the model would be worse than in the case of an author writing on one topic throughout their works.}

Since our main interest was in investigating an authorship analysis system's performance when applied to two different documents by the same author (such as Arata Osada), samples where both snippets originate from the same document were not indispensable for the training of our classifier.
At the same time, we expected that reserving more capacity in the training set for the other two types of positive samples may result in improved accuracy.
To verify that hypothesis, in an additional experiment we trained a model on data without same-document samples.

In order to minimize the potential effect that the characteristics of a particular random sample extracted from our data might have on the experiment results,
we generated all data sets three times,
each time with a different author split.
In Section~\ref{sec:results}, we report the results calculated from the sum of predictions made by models trained and tested on all three variants.

Tables~\ref{tab:data_stats_5cats} and~\ref{tab:data_stats_4cats} show statistics of data sets used in both experiments.
A sample data point from the training set is shown in Table~\ref{tab:data_sample}.
During training, the model was only presented with the last two fields: ``Paragraph1+2'' and ``Label''.

\begin{table}[htb]
\centering
\begin{tabular}{rccccccc}
\toprule
& \rotatebox[origin=r]{270}{\footnotesize{Aozora train.}} & \rotatebox[origin=r]{270}{\footnotesize{Aozora dev.}} & \rotatebox[origin=r]{270}{\footnotesize{Aozora test}} & \rotatebox[origin=r]{270}{\footnotesize{Y. Fukuzawa}} & \rotatebox[origin=r]{270}{\footnotesize{K. Izumi}} & \rotatebox[origin=r]{270}{\footnotesize{A. Oka}} & \rotatebox[origin=r]{270}{\footnotesize{A. Osada}} \\
\midrule
Authors & 356 & 22 & 34 & 35 & 35 & 35 & 35 \\
\hline
Documents (mean) & 4395 & 343 & 312 & 224 & 242 & 235 & 252 \\
\hline
& \multicolumn{7}{c}{\footnotesize{Same author \& document}} \\
& 6,000 & 400 & 600 & 600 & 600 & 600 & 600 \\
\cline{2-8}
& \multicolumn{7}{c}{\footnotesize{Same author, diff. documents, dist. <= 10 y.}} \\
& 6,000 & 400 & 600 & 600 & 600 & 600 & 600 \\
\cline{2-8}
\multirow{2}{*}{Data points} & \multicolumn{7}{c}{\footnotesize{Same author, diff. documents, dist. > 10 y.}} \\
& 6,000 & 400 & 600 & 600 & 600 & 600 & 600 \\
\cline{2-8}
& \multicolumn{7}{c}{\footnotesize{Different authors, distance <= 10 years}} \\
& 9,000 & 600 & 900 & 900 & 900 & 900 & 900 \\
\cline{2-8}
& \multicolumn{7}{c}{\footnotesize{Different authors, distance > 10 years}} \\
& 9,000 & 600 & 900 & 900 & 900 & 900 & 900 \\
\bottomrule
\end{tabular}
\caption{Statistics of data sets used in the experiment with all 5 types of samples. Since the numbers of documents in each of the 3 variants differ, we report the average values.}
\label{tab:data_stats_5cats}
\end{table}

\begin{table}[htb]
\centering
\begin{tabular}{rccccccc}
\toprule
& \rotatebox[origin=r]{270}{\footnotesize{Aozora train.}} & \rotatebox[origin=r]{270}{\footnotesize{Aozora dev.}} & \rotatebox[origin=r]{270}{\footnotesize{Aozora test}} & \rotatebox[origin=r]{270}{\footnotesize{Y. Fukuzawa}} & \rotatebox[origin=r]{270}{\footnotesize{K. Izumi}} & \rotatebox[origin=r]{270}{\footnotesize{A. Oka}} & \rotatebox[origin=r]{270}{\footnotesize{A. Osada}} \\
\midrule
Authors & 356 & 22 & 34 & 35 & 35 & 35 & 35 \\
\hline
Documents (mean) & 4597 & 365 & 323 & 219 & 248 & 242 & 255 \\
\hline
& \multicolumn{7}{c}{\footnotesize{Same author, diff. documents, dist. <= 10 y.}} \\
& 9,000 & 600 & 900 & 900 & 900 & 900 & 900 \\
\cline{2-8}
& \multicolumn{7}{c}{\footnotesize{Same author, diff. documents, dist. > 10 y.}} \\
\multirow{2}{*}{Data points} & 9,000 & 600 & 900 & 900 & 900 & 900 & 900 \\
\cline{2-8}
& \multicolumn{7}{c}{\footnotesize{Different authors, distance <= 10 years}} \\
& 9,000 & 600 & 900 & 900 & 900 & 900 & 900 \\
\cline{2-8}
& \multicolumn{7}{c}{\footnotesize{Different authors, distance > 10 years}} \\
& 9,000 & 600 & 900 & 900 & 900 & 900 & 900 \\
\bottomrule
\end{tabular}
\caption{Statistics of data sets used in the experiment without same-document samples. Since the numbers of documents in each of the 3 variants differ, we report the average values.}
\label{tab:data_stats_4cats}
\end{table}

\begin{table}[htbp]
\centering
\begin{tabularx}{\textwidth}{rX}
\toprule
Author1 & 000525 \\
Year1 & 1933 \\
Document1 & 43239 \\
Paragraph1 & \scriptsize{\begin{CJK}{UTF8}{ipxm}「先づ今日の時勢よりお話申しますと、世人の社会を見ることが、簡単過ぎて居ると思ひます。 [...] また政治の領分は、社会的事物の大なる者だが、社会全体から見ればその一部分である。されば仮令政治の弊害全部を破りました所で、社会の一部が良くならうが、全部は良くなりませぬ。\end{CJK}} \\

Author2 & 000525 \\
Year2 & 1904 \\
Document2 & 3507 \\
Paragraph2 & \scriptsize{\begin{CJK}{UTF8}{ipxm}「左様です、彼は決して嫉妬などの為めに凶行に出でたのではありません、 [...] 又た其の高潔なる愛情の手に倒れたと云ふことは、女性としての満足なる生涯では無いでせうか」\end{CJK}} \\

Paragraph1+2 & \scriptsize{\begin{CJK}{UTF8}{ipxm}「先づ今日の時勢よりお話申しますと、世人の社会を見ることが、簡単過ぎて居ると思ひます。欧羅巴の人々も過去の歴史に於て、矢張りこの過ちを重ねて居ります。それは何かと言ふに、政治上目に見える弊害があれば、その弊だけを止めれば、社会は大層良くなると思うて、尽力して弊を除きましても、社会は思つた程良くならない。こゝに於て大失望して大騒動となる。我国に於きましても、政治上に種々なる弊害があるから、欧羅巴の様に改革したならば、定めし黄金世界になるであらうと思つて居る者があるが、これは余りに政治に重きを置き過ぎたものと、私は考へます。社会を組立てるには色々道具がある。家族であるとか、国民の教育であるとか、これを集めて社会が出来て居る。この社会的原素の中に政治と云ふものがある。また政治の領分は \textbf{[SEP]} 「左様です、彼は決して嫉妬などの為めに凶行に出でたのではありません、――必竟、自分の最愛の妻――仮令結婚はしないにせよ――を、姦淫の罪悪から救はねばならぬと云ふのが、彼の最終の決心であつたのです、彼の此の愛情は独り婦人に対してのみで無いのです、彼が平生、職業に対し、友人に対し、事業に対する観念が皆な其れでした、成程、其の小米と云ふ婦人も、今ま貴女の（と花吉を一瞥しつ）仰つしやる通り実に気の毒でした、然かし彼女が彼の如くして生きて居たからとて、一日と雖も、一時間と雖も、幸福と云ふ感覚を有つことは無かつたでせう、兼吉が執つた婦人に対する最後の手段は、無論正道をば外れてたでせう、が、生まれて此の如き清浄な男児の心を得、又た其の高潔なる愛情の手に倒れた\end{CJK}} \\

Label & 1 \\
\bottomrule
\end{tabularx}
\caption{Sample record from the training data \hl{Text omissions are indicated by ellipses in square brackets (``[...]''). Translations are provided in Table}~\ref{tab:data_sample_en}.}
\label{tab:data_sample}
\end{table}

\begin{table}[htbp]
\centering
\begin{tabularx}{\textwidth}{rX}
\toprule
Author1 & 000525 \\
Year1 & 1933 \\
Document1 & 43239 \\
Paragraph1 & \scriptsize{\hl{First of all, looking at the world as it is nowadays,  I think it is too easy to observe the human society. [...] Also, when it comes to the wold of politics,  great persons do exist in society, but are only a part of it. As so, when all the harmful effects of provisional politics have been conquered, part of society will improve, but not all of it.}} \\

Author2 & 000525 \\
Year2 & 1904 \\
Document2 & 3507 \\
Paragraph2 & \scriptsize{\hl{That's right, he never went out of his way because of jealousy, [...] and surrendering to one's noble affection is a satisfying life for a woman. Don't you think?}} \\

Paragraph1+2 & \scriptsize{\hl{First of all, looking at the world as it is nowadays, I think it is too easy to observe the human society. The people of Europe have also made this mistake multiple times in the past. What is it? Well, it is the thought that if there are any  visible harmful effects of policies, stopping only those harmful effects will make the society much better; however, even if we do our best to get rid of what is wrong, society will not improve as much as we think. It provokes great disappointment and a big uproar. Even in Japan, there are various harmful phenomena in politics, so some people think that if they introduce reforms like the ones in Europe, we will come to live in a golden world,
but this for me is putting too much emphasis on politics. There are various tools that can be used to build a society. A society is formed by collecting these, such as ``family'' and ``education of the people''. There is also something called politics as one of the elements of society. Also, the territory of politics is} \textbf{[SEP]} \hl{That's right, he never went out of his way because of jealousy. It was his final decision to save his beloved wife -- even if she didn't marry him -- from the sin of adultery. His love isn't just for a single woman. His approach is identical when it comes to his life, his profession, his friends, his business. 
I see, all was really just as that lady, Shomei (and she took a glance at Hanayoshi), said. It was really a pity. However, because she lived like him, she did not have a sense of happiness for even a day or even an hour. The last resort for women is, of course, out of the ordinary, but he was born and with a pure boy's heart like this, and fell into the hands of his noble affection.}} \\

Label & 1 \\
\bottomrule
\end{tabularx}
\caption{Sample record from the training data (translated to English). Text omissions are indicated by ellipses in square brackets (``[...]'').}
\label{tab:data_sample_en}
\end{table}

\subsection{Model training}

To train our classification system, we fine-tuned the Japanese BERT model on the Aozora training set for one epoch (in preliminary experiments we fine-tuned for up to 5 epochs, but no further improvements were observed on the development set).
We used a learning rate of 3e-6 and batch size of 16.
Our classifier was implemented with the Flair library\footnote{\url{https://github.com/flairNLP/flair}}~\cite{akbik2018coling}.
\hl{Unlike the bulk of previous studies} (e.g., \cite{koppel2014,Hrlimann2015GLADGL,Boenninghoff2019ExplainableAV}), \hl{we did not employ an explicit similarity metric, such as cosine or Euclidean distance, to compare pairs of documents in a feature space. Instead, we used the transformer model to obtain (by the means of the special \textit{[CLS]} token) a joint representation of both documents in each verification problem and trained a linear layer on top of it to directly predict the label.}

As explained in the previous section, all data sets were created in three different permutations.
Furthermore, since we noticed a fair amount of variability in the validation results between multiple training runs on the same data, we repeated the training process 5 times on each variant of the data.
As a result, we trained and tested a total of 15 models -- in Section~\ref{sec:results}, we report the results calculated from the sum of predictions made by all of them.

\subsection{Baseline method}

\hl{In order to verify the effectiveness of our authorship verification method, we performed a baseline experiment using the Impostors Method proposed by Koppel and Winter} \cite{koppel2014}\footnote{Specifically, we used the implementation by Mike Kestemont (see \cite{Kestemont2016AuthenticatingTW}), released in \url{https://github.com/mikekestemont/ruzicka}.}, which has demonstrated strong performance on multiple data sets \cite{Kestemont2016AuthenticatingTW,Halvani2019UnaryAB}.
It is an example of an \textit{extrinsic} authorship verification method where additional documents are used as negative samples \cite{PothaStamatatos2018_intrinsic}.
Specifically, it compares a randomly selected subset of features observed in the disputed document with one or more documents by the target author and a set of ``impostor'' documents by other authors.
The procedure is repeated multiple times, each time using a different subset of features and impostor documents.
If the proportion of iterations in which the candidate author's document was closer to the disputed text than any of the distractor documents exceeds a certain threshold \hl{(in our case, 50\%)}, then both documents are attributed to the same author.

In our experiments, \texttt{tf-idf} \hl{values of the 50,000 most frequent character 2-grams in the training set were used as features.
We repeated the feature randomization procedure 100 times for each verification problem, sampling 90\% of the available features in each iteration.
For each test sample, we selected a random subset of 100 potential impostor documents from the training set, of which 5 were randomly sampled and actually used in each iteration.
The distance between pairs of documents was measured using the} \textit{minmax} metric.
\hl{All the above hyperparameters were tuned on the development set.}
We treated the second document (``Paragraph2'' -- see Table~\ref{tab:data_sample}) in each test sample as the anonymous text and compared it to impostor documents.
All paragraphs were truncated to the same length as in the experiments using BERT.

\section{Experiment results}
\label{sec:results}

\begin{figure}
    \centering
    \includegraphics[width=\linewidth]{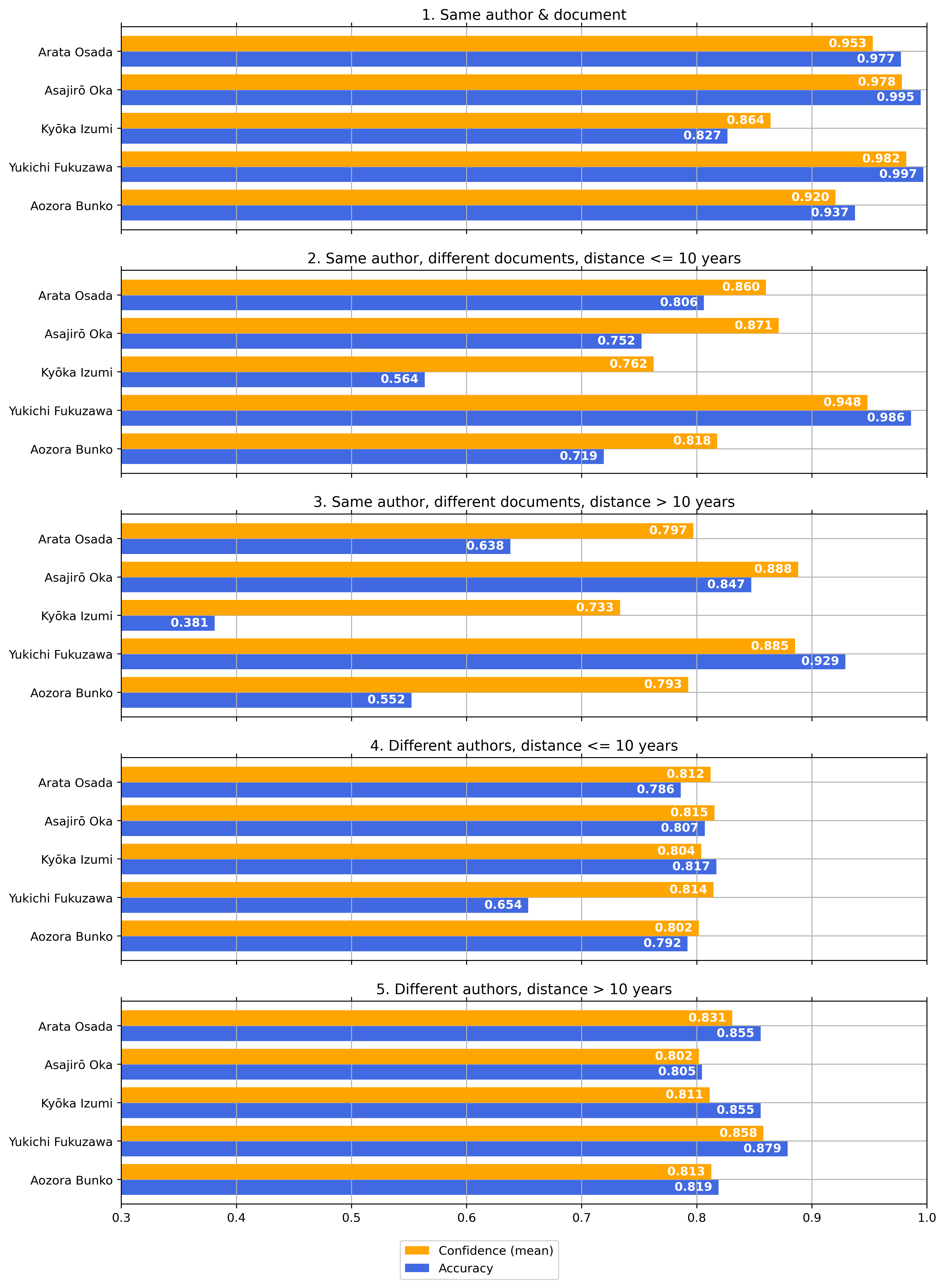}
    \caption{Results by category of test samples (BERT).}
    \label{fig:bar_graph_5_cats}
\end{figure}

\begin{figure}
    \centering
    \includegraphics[width=0.925\linewidth]{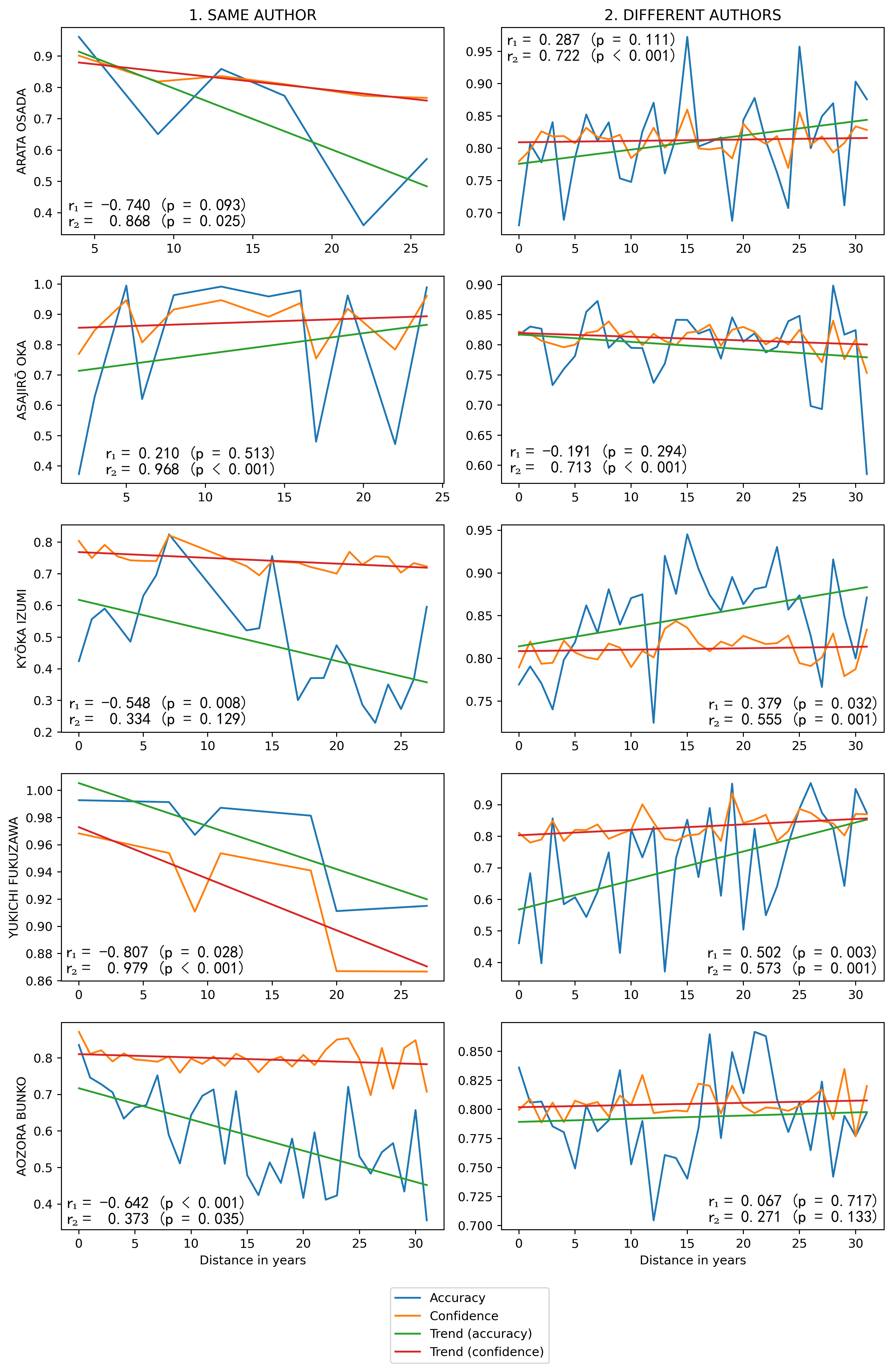}
    \caption{Relation between the distance in years
	between two documents and system's performance (BERT). We also report the values of Pearson correlation coefficient between the distance in years and accuracy ($r_1$) and between confidence and accuracy ($r_2$), as well as their respective p-values.}
    \label{fig:dist2acc}
\end{figure}

The results of the main experiment \hl{using BERT} are summarized in Table~\ref{tab:rslts_with_baseline} \hl{(in comparison with the baseline method)} and Figures~\ref{fig:bar_graph_5_cats} and~\ref{fig:dist2acc}.
In Figure~\ref{fig:bar_graph_5_cats}, the bottom row of each graph represents the results on the test set made based on Aozora Bunko, while the four other rows represent the results on the samples based on works of Arata Osada and the three authors from the comparison sample.
The blue bar represents the accuracy with which the model was able to identify authorship, while the orange bar represents the confidence with which the choice was performed (i.e., softmax probability of the predicted class).

\begin{table}[htb]
\centering
\begin{tabular}{rccccccc}
\toprule
& & \multicolumn{3}{c}{\textbf{Impostors}} & \multicolumn{3}{c}{\textbf{BERT}} \\
\textbf{\footnotesize{Test set}} & \textbf{\footnotesize{Class}} & \textbf{\footnotesize{Prec.}} & \textbf{\footnotesize{Rec.}} & \textbf{\footnotesize{F\textsubscript{1}}} & \textbf{\footnotesize{Prec.}} & \textbf{\footnotesize{Rec.}} & \textbf{\footnotesize{F\textsubscript{1}}}\\
\midrule
Arata & 1 & \textbf{0.818} & 0.697 & 0.753 & \textbf{0.818} & \textbf{0.807} & \textbf{0.813} \\
\cline{2-8}
Osada & 0 & 0.736 & \textbf{0.845} & 0.787 & \textbf{0.810} & 0.821 & \textbf{0.815} \\
\hline
Asajirō & 1 & \textbf{0.824} & 0.810 & 0.817 & 0.817 & \textbf{0.865} & \textbf{0.840} \\
\cline{2-8}
Oka & 0 & 0.813 & \textbf{0.827} & 0.820 & \textbf{0.856} & 0.806 & \textbf{0.830} \\
\hline
Kyōka & 1 & \textbf{0.787} & 0.422 & 0.549 & 0.783 & \textbf{0.590} & \textbf{0.673} \\
\cline{2-8}
Izumi & 0 & 0.605 & \textbf{0.886} & 0.719 & \textbf{0.671} & 0.836 & \textbf{0.745} \\
\hline
Yukichi & 1 & \textbf{0.820} & \textbf{0.985} & \textbf{0.895} & 0.806 & 0.971 & 0.881 \\
\cline{2-8}
Fukuzawa & 0 & \textbf{0.981} & 0.\textbf{784} & \textbf{0.872} & 0.963 & 0.766 & 0.853 \\
\hline
Aozora & 1 & \textbf{0.817} & 0.605 & 0.695 & 0.791 & \textbf{0.736} & \textbf{0.763} \\
\cline{2-8}
Bunko & 0 & 0.687 & \textbf{0.865} & 0.765 & \textbf{0.753} & 0.805 & \textbf{0.778} \\
\hline
\multirow{2}{*}{OVERALL} & 1 & \textbf{0.816} & 0.704 & 0.756 & 0.804 & \textbf{0.794} & \textbf{0.799} \\
\cline{2-8}
& 0 & 0.740 & \textbf{0.841} & 0.787 & \textbf{0.796} & 0.807 & \textbf{0.802} \\
\bottomrule
\end{tabular}
\caption{Experiment results (best results in bold).}
\label{tab:rslts_with_baseline}
\end{table}

In the first graph (Fig.~\ref{fig:bar_graph_5_cats}.1), we present the results of authorship verification on a combination of two fragments from the same document.
In this case, the system performed well above the average, with accuracy exceeding 90\% on all but one test set and 99\% on two of them. This means that, apart from being able to capture the relationship between adjacent sentences (as in the next sentence prediction objective), BERT is also capable of discerning similarities in content between distant samples from the same work.

The second graph (Fig.~\ref{fig:bar_graph_5_cats}.2) demonstrates how the model performed when presented with paragraphs of texts written by the same author, but coming from different books, written in the space of less than or 10 years.
The third graph (Fig.~\ref{fig:bar_graph_5_cats}.3) is where results are given for a setting with paragraphs of texts written by the same author, but coming from different books, written in the space of more than 10 years.
In the first case, a drop in accuracy of 18.1\% on average was observed, and increasing the distance in years resulted in further drop by 9.6\%.

In all three categories of positive samples, results for Kyōka Izumi were the lowest, which is consistent with our hypothesis that a diversified sample such as excerpts of novels with different plots and subjects and thus diverse content would affect the performance of the model.
It is also confirmed by the fact that the results on the Aozora Bunko set -- consisting largely of fiction works -- were the second worst.
Surprisingly this tendency is also visible in the case of samples created from a single document, implying that novels are more diverse in terms of content also within a single document and as such represent more of a challenge for authorship analysis systems.

\begin{figure}
    \centering
    \includegraphics[width=\linewidth]{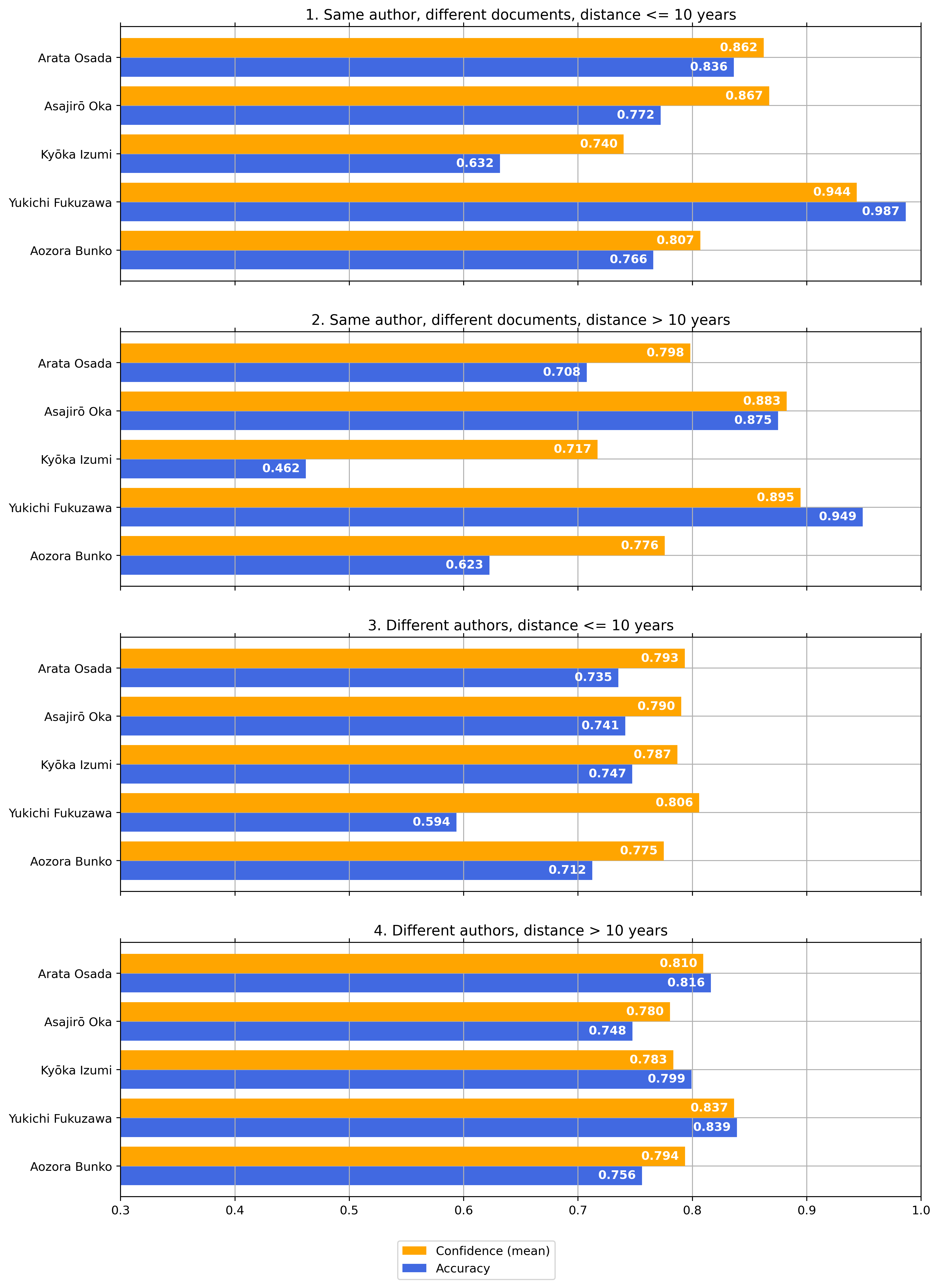}
    \caption{Results for the model trained on data without same-author samples (BERT).}
    \label{fig:bar_graph_4_cats}
\end{figure}

Table~\ref{tab:rslts_4_cats} and Figure~\ref{fig:bar_graph_4_cats} present the results yielded by the \hl{BERT-based} classifier trained on data without same-document samples and a comparison with the model trained on all five types of samples.
While it did correctly detect same authorship for a greater number of samples from the remaining two categories of positive samples (presumably due to increased capacity in the training data), it was at the cost of noticeably lower F-score for different-author samples.
This suggests that training samples where both fragments of text originate from the same document are beneficial in the modeling of other categories, as well.

\begin{table}[htb]
\centering
\begin{tabular}{rccccccc}
\toprule
& & \multicolumn{6}{c}{\textbf{\footnotesize{Model trained with same-document samples:}}} \\
& & \multicolumn{3}{c}{\textbf{NO}} & \multicolumn{3}{c}{\textbf{YES}} \\
\textbf{\footnotesize{Test set}} & \textbf{\footnotesize{Class}} & \textbf{\footnotesize{Prec.}} & \textbf{\footnotesize{Rec.}} & \textbf{\footnotesize{F\textsubscript{1}}} & \textbf{\footnotesize{Prec.}} & \textbf{\footnotesize{Rec.}} & \textbf{\footnotesize{F\textsubscript{1}}}\\
\midrule
Arata & 1 & 0.775 & \textbf{0.772} & \textbf{0.773} & \textbf{0.805} & 0.719 & 0.760 \\
\cline{2-8}
Osada & 0 & \textbf{0.773} & 0.776 & 0.774 & 0.746 & \textbf{0.826} & \textbf{0.784} \\
\hline
Asajirō & 1 & 0.763 & \textbf{0.824} & 0.792 & \textbf{0.807} & 0.796 & \textbf{0.802} \\
\cline{2-8}
Oka & 0 & \textbf{0.808} & 0.744 & 0.775 & 0.799 & \textbf{0.810} & \textbf{0.804} \\
\hline
Kyōka & 1 & 0.707 & \textbf{0.547} & \textbf{0.617} & \textbf{0.738} & 0.470 & 0.574 \\
\cline{2-8}
Izumi & 0 & \textbf{0.631} & 0.773 & 0.695 & 0.611 & \textbf{0.834} & \textbf{0.705} \\
\hline
Yukichi & 1 & 0.773 & \textbf{0.968} & 0.860 & \textbf{0.808} & 0.956 & \textbf{0.876} \\
\cline{2-8}
Fukuzawa & 0 & \textbf{0.957} & 0.716 & 0.819 & 0.947 & \textbf{0.773} & \textbf{0.851} \\
\hline
Aozora & 1 & 0.723 & \textbf{0.694} & \textbf{0.708} & \textbf{0.765} & 0.635 & 0.694 \\
\cline{2-8}
Bunko & 0 & \textbf{0.706} & 0.734 & 0.720 & 0.688 & \textbf{0.805} & \textbf{0.742} \\
\hline
\multirow{2}{*}{OVERALL} & 1 & 0.752 & \textbf{0.761} & \textbf{0.756} & \textbf{0.790} & 0.715 & 0.751 \\
\cline{2-8}
& 0 & \textbf{0.758} & 0.749 & 0.753 & 0.740 & \textbf{0.810} & \textbf{0.773} \\
\bottomrule
\end{tabular}
\caption{Comparison of the results achieved by BERT-based models trained with and without same-document samples, on test data without such samples (best results in bold).}
\label{tab:rslts_4_cats}
\end{table}

Based on the experiment, what answer should be given to our main research question: did the change in Osada's ethical values reflect itself in the predictions yielded by the authorship analysis system?
While -- as shown in Fig.~\ref{fig:bar_graph_5_cats}.2 and~\ref{fig:bar_graph_5_cats}.3 and Fig.~\ref{fig:dist2acc} (first column) -- there was a significant drop in accuracy as the distance in years between two documents increased,
the same was also true for other authors (with the exception of Asajirō Oka) and the Aozora Bunko data set.
The fact that the results on test data based on fiction works (i.e., Kyōka Izumi's novels and the Aozora Bunko data set) were consistently worse than those for Osada's books, seems to indicate that a change in the author's opinions is less relevant to the model than a change in topic (such as between two novels by the same author, with different characters and/or settings).
Furthermore, we did not observe a clear-cut difference between the results on test samples composed exclusively of fragments from pre-war or post-war books and those combining texts from both periods: the accuracy for samples including one paragraph from each of Osada's two pre-war books (i.e., those where the distance in years is 9 -- see Fig.~\ref{fig:dist2acc}) was lower than for samples combining the \textit{Shinkyōiku no kōsō -- Amerika no bunka-kyōiku wo hihan shite} with fragments from either of the post-war books (i.e., those where the time lapse is 13 or 17 years).

On the other hand, in terms of accuracy for samples comprising paragraphs from distant documents, the results on Osada test set were substantially lower than in the case of other non-fiction writers (i.e., Yukichi Fukuzawa and Asajirō Oka).
Interesting conclusions can also be drawn from observation of the changes in probabilities assigned by the classifier to its predictions.
While in the case of Asajirō Oka and Yukichi Fukuzawa,
\hl{nearly perfect correlation ($r > 0.95$) is observed}
between the confidence scores and accuracy, on the remaining three data sets -- including Osada test set -- there is a widening gap between the two values as the time lapse increases (see Fig.~\ref{fig:dist2acc}), which means that the model was confident that texts written by the same author were actually written by two different people.
This leads us to believe that a change in opinions as expressed in writing can affect authorship analysis. If before and after experiencing a World War the opinions that Arata Osada formulated in his writings changed enough for the model to encounter difficulties while trying to detect single authorship, this might also mean that the impact of historical events, such as a war, on a person's ethical outlook is significant enough to make them express themselves in writing like a different person or in other words ``make them a different person''.

The fourth (Fig.~\ref{fig:bar_graph_5_cats}.4) and fifth (Fig.~\ref{fig:bar_graph_5_cats}.5) graph and the second column of Fig.~\ref{fig:dist2acc} illustrate the results when the classifier was presented with two fragments of text written by different authors.
Contrary to same-author samples, in this case the performance tends to increase with the distance between the dates of publication, since the style of writing and language evolves over time, making it easier for the model to make a distinction between authors and come to the conclusion that samples provided are authored by two different people.

\begin{figure}
    \centering
    \includegraphics[width=0.925\linewidth]{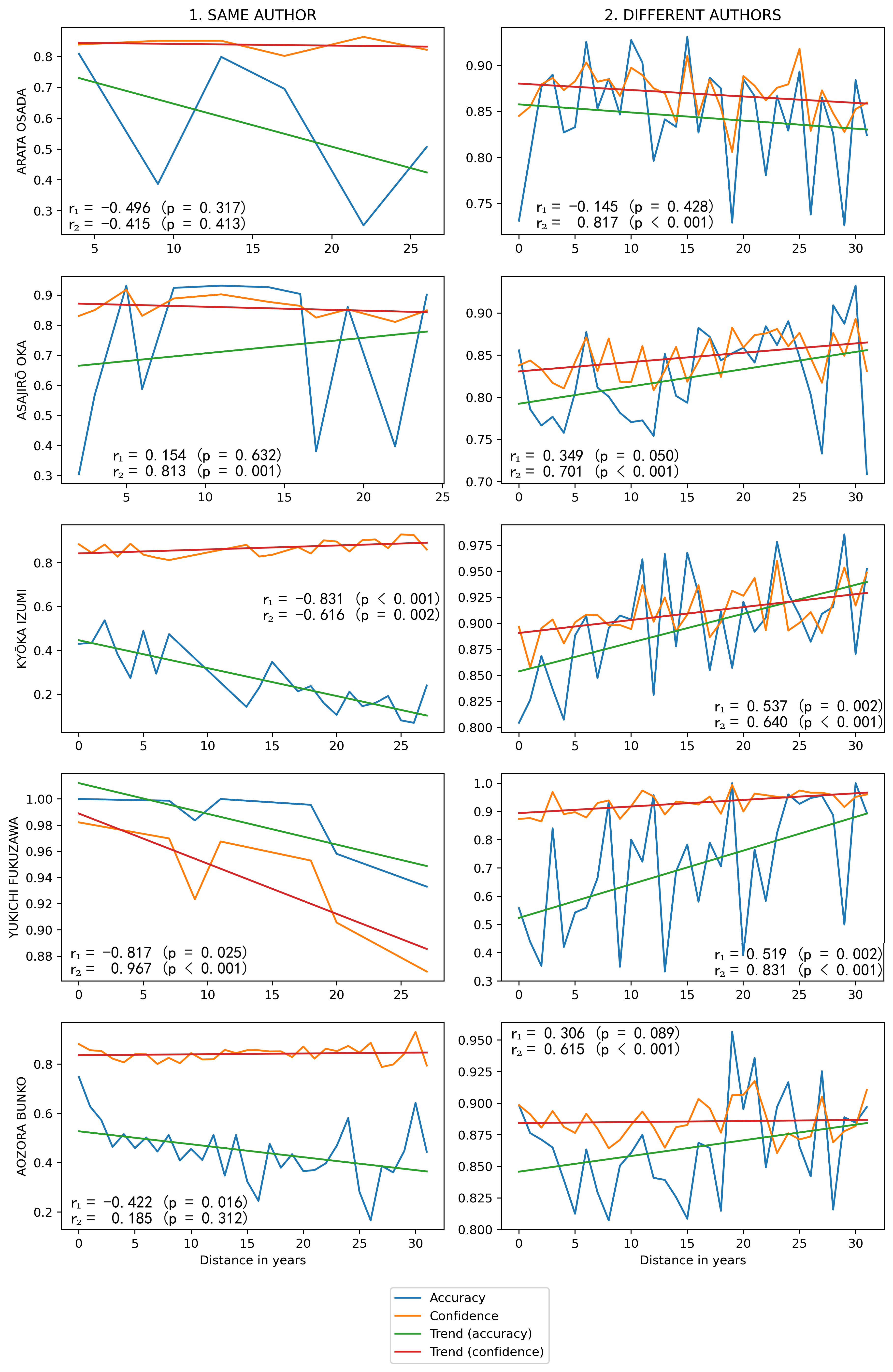}
    \caption{Relation between the distance in years
	between two documents and system's performance (Impostors). We also report the values of Pearson correlation coefficient between the distance in years and accuracy ($r_1$) and between confidence and accuracy ($r_2$), as well as their respective p-values. }
    \label{fig:dist2accBaseline}
\end{figure}

\hl{The baseline method exhibited a strong bias towards the negative class and yielded lower F-scores than BERT for 4 out of 5 test sets, the only exception being the works of Yukichi Fukuzawa} (Table~\ref{tab:rslts_with_baseline}).
\hl{McNemar's test indicated that there was a statistically significant difference between the results obtained by both classifiers} ($p < 0.001$).
\hl{Furthermore, confidence of the predictions made by BERT when tested on samples by the same author remained a better indicator of their accuracy -- in the case of the baseline system, the positive correlation between both factors\footnote{In the case of the Impostors Method-based classifier, confidence refers to the proportion of feature sets for which the target author's document (i.e., the first document in the given test sample) was more similar to the disputed document than any of the impostor documents, if it exceeds 50\%, or the complement of that proportion, otherwise.} was weaker or absent} (see Figures~\ref{fig:dist2acc} and~\ref{fig:dist2accBaseline}).
\hl{This suggests that a transformer-based authorship verification system is harder to ``fool'' by changes in topics and/or opinions expressed an author than traditional methods relying on distance metrics.}

\section {Discussion}
\label{sec:discussion}

Currently developed solutions in the area of automatic authorship analysis are based on the assumption that there are certain stable, unchanging factors or features to the content an individual generates, on which the whole process of identification is founded.
However the results of
our experiments, particularly the extent to which a classification system's accuracy in detecting same authorship deteriorated when presented with texts written by the same author over a longer time span, show the need to add the aspect of time-related changes into consideration when performing authorship analysis.

This leaves us with the question of whether there actually exists any trait that remains consistent throughout one’s writing, and if not -- what factors (content? style?) specifically change in a person’s writing over time, in what measure, how to quantify this change, how off can you be in predicting a writer’s evolution and whether there exists a pattern of behavioral change observable through someone‘s writing. Tracing, evaluating and predicting such changes might be useful especially in the case of practical solutions in the field of forensics (terrorism countermeasures) or online bullying detection. 

Another question is, from the point of view of the ethical evolution of an individual, what exactly constitutes a formative experience major enough to provoke an observable shift in opinions. In the case of the subject of our research, Arata Osada, the change in opinions can be traced back to his traumatic experience of the Second World War.  In the case of different authors, who have not been subjected to major external historical stimuli, however, the question of whether a change is also noticeable due to natural evolution of opinions juxtaposed with the passage of time, and how to quantify this evolution, remains open.

\section{Conclusions and future work}
\label{sec:conclusions}

In this research we wanted to find out whether a model created for the task of same authorship detection would correctly attribute works from different points in time to the same author and with what accuracy.
In particular, we were interested to see if the system experiences any additional difficulty in single authorship identification when presented with two texts by a person whose opinions and/or ethical values changed in the intervening period between writing
them (e.g., in the case of the main object of this research, Arata Osada, works written pre- and post-World War II) -- which would mean that the impact of historical events on a person’s ethical outlook and the content (books, articles, etc.) he or she produces, is significant enough that it can be quantified.

We conducted this research through performing a binary authorship analysis task using a text classifier based on a pre-trained transformer model, fine-tuned on randomly sampled pairs of text snippets from a repository of Japanese literary works (Aozora Library).
The system was then tested on five different data sets, including a test set generated on the basis of four books authored by Arata Osada, a Japanese educator and specialist in the history of education (of which two were books written before the Second World War and another two, in the 1950s) and a three comparison sets composed of works by three other Japanese authors: Yukichi Fukuzawa, Kyōka Izumi and Asajirō Oka.
The task of the classifier was to identify whether there is one or many authors of the texts presented to it.
\hl{In order to validate our approach, we compared it with one of state-of-the-art authorship verification methods (namely, the Impostors Method) and found the transformer-based classification system to yield better performance on most data sets.}

Upon performing the experiment, we found out that there is a strong negative correlation between the amount of time elapsed between the publication of two documents by a single person and the system's accuracy in detecting their common authorship, with the drop in performance being the highest for fiction books (i.e., novels), due to higher contentual diversity.
In the group of non-fiction writers, the model had the lowest accuracy when presented with two fragments of Arata Osada's texts, one of them originating from before and one from after the Second World War.
This and the observation that the classifier maintained relatively high confidence in its incorrect judgements, indicate that a major shift in one’s opinions as reflected in writing -- although it might have less impact on the classification process than a change of topic -- is often enough to convince the classification model that the authors are two different people, and by consequence that historical events such as a war can change the way an author expresses his or her opinions in writing beyond the point of recognition for an authorship verification solution.

As a next step in our research, we are planning to extend the analysis presented in this paper by considering and comparing the effect of different aspects relevant to the process of authorship analysis, such as style (linguistic features) and contentual characteristics (semantics and topics).
As another direction of further research, we would like to analyze emotional attitudes towards different ethically relevant concepts in the works of the same author (sentiment analysis).

We also plan to define a list of text features that the model should take into consideration in order for it to be applied to other tasks such as authorship analysis in online harassment scenarios and broaden this set of features to make it adaptable to use with an artificial companion.

\bibliography{article}

\begin{thebibliography}{10}
\expandafter\ifx\csname url\endcsname\relax
  \def\url#1{\texttt{#1}}\fi
\expandafter\ifx\csname urlprefix\endcsname\relax\def\urlprefix{URL }\fi
\expandafter\ifx\csname href\endcsname\relax
  \def\href#1#2{#2} \def\path#1{#1}\fi

\bibitem{Mendenhall237}
T.~C. Mendenhall,
  \href{https://science.sciencemag.org/content/ns-9/214S/237}{The
  characteristic curves of composition}, Science ns-9~(214S) (1887) 237--246.
\newblock \href
  {http://arxiv.org/abs/https://science.sciencemag.org/content/ns-9/214S/237.full.pdf}
  {\path{arXiv:https://science.sciencemag.org/content/ns-9/214S/237.full.pdf}},
  \href {https://doi.org/10.1126/science.ns-9.214S.237}
  {\path{doi:10.1126/science.ns-9.214S.237}}.
\newline\urlprefix\url{https://science.sciencemag.org/content/ns-9/214S/237}

\bibitem{Chaski2013BestPA}
C.~Chaski, Best practices and admissibility of forensic author identification,
  Journal of law and policy 21 (2013) 5.

\bibitem{frommholz2017textual}
I.~Frommholz, K.~M, M.~Potthast, Z.~Ghasem, M.~Shukla, E.~Short,
  \href{https://books.google.co.jp/books?id=A\_wXtAEACAAJ}{On Textual Analysis
  and Machine Learning for Cyberstalking Detection}, Bauhaus-Universit{\"a}t
  Weimar, 2017.
\newline\urlprefix\url{https://books.google.co.jp/books?id=A\_wXtAEACAAJ}

\bibitem{Bovet}
A.~Bovet, H.~Makse, Influence of fake news in twitter during the 2016 us
  presidential election, Nature Communications 10 (01 2019).
\newblock \href {https://doi.org/10.1038/s41467-018-07761-2}
  {\path{doi:10.1038/s41467-018-07761-2}}.

\bibitem{Juola2015IndustrialUF}
P.~Juola, Industrial uses for authorship analysis, 2015.

\bibitem{Sari2018TopicOS}
Y.~Sari, M.~Stevenson, A.~Vlachos, Topic or style? exploring the most useful
  features for authorship attribution, in: COLING, 2018.

\bibitem{nieuwazny}
J.~Nieuwazny, K.~Nowakowski, M.~Ptaszynski, R.~Rzepka, F.~Masui, K.~Araki, Does
  change in ethical education influence core moral values? towards
  culture-aware morality model, Cognitive Systems Research (2020) in press.

\bibitem{zheng}
R.~Zheng, J.~Li, H.-c. Chen, Z.~Huang, A framework for authorship
  identification of online messages: Writing-style features and classification
  techniques, JASIST 57 (2006) 378--393.
\newblock \href {https://doi.org/10.1002/asi.20316}
  {\path{doi:10.1002/asi.20316}}.

\bibitem{mosteller}
F.~Mosteller, D.~L. Wallace,
  \href{http://www.jstor.org/stable/2283270}{Inference in an authorship
  problem}, Journal of the American Statistical Association 58~(302) (1963)
  275--309.
\newline\urlprefix\url{http://www.jstor.org/stable/2283270}

\bibitem{article}
D.~Holmes, J.~Kardos, Who was the author? an introduction to stylometry, CHANCE
  16 (03 2003).
\newblock \href {https://doi.org/10.1080/09332480.2003.10554842}
  {\path{doi:10.1080/09332480.2003.10554842}}.

\bibitem{stamatatos}
E.~Stamatatos, A survey of modern authorship attribution methods, JASIST 60
  (2009) 538--556.
\newblock \href {https://doi.org/10.1002/asi.21001}
  {\path{doi:10.1002/asi.21001}}.

\bibitem{madigan}
D.~Madigan, A.~Genkin, D.~Lewis, S.~Argamon, D.~Fradkin, L.~Ye, D.~Consulting,
  Author identification on the large scale (01 2005).

\bibitem{Potha2014}
N.~Potha, E.~Stamatatos, A profile-based method for authorship verification,
  in: A.~Likas, K.~Blekas, D.~Kalles (Eds.), Artificial Intelligence: Methods
  and Applications, Springer International Publishing, Cham, 2014, pp.
  313--326.

\bibitem{koppel2014}
M.~Koppel, Y.~Winter,
  \href{https://asistdl.onlinelibrary.wiley.com/doi/abs/10.1002/asi.22954}{Determining
  if two documents are written by the same author}, Journal of the Association
  for Information Science and Technology 65~(1) (2014) 178--187.
\newblock \href
  {http://arxiv.org/abs/https://asistdl.onlinelibrary.wiley.com/doi/pdf/10.1002/asi.22954}
  {\path{arXiv:https://asistdl.onlinelibrary.wiley.com/doi/pdf/10.1002/asi.22954}},
  \href {https://doi.org/10.1002/asi.22954} {\path{doi:10.1002/asi.22954}}.
\newline\urlprefix\url{https://asistdl.onlinelibrary.wiley.com/doi/abs/10.1002/asi.22954}

\bibitem{Hrlimann2015GLADGL}
M.~H{\"u}rlimann, B.~Weck, E.~van~den Berg, M.~Nissim, {GLAD: Groningen
  Lightweight Authorship Detection -- Notebook for PAN at CLEF 2015}, 2015.

\bibitem{Boenninghoff2019ExplainableAV}
B.~T. Boenninghoff, S.~Hessler, D.~Kolossa, R.~M. Nickel, Explainable
  authorship verification in social media via attention-based similarity
  learning, 2019 IEEE International Conference on Big Data (Big Data) (2019)
  36--45.

\bibitem{Kestemont2018OverviewOT}
M.~Kestemont, M.~Tschuggnall, E.~Stamatatos, W.~Daelemans, G.~Specht, B.~Stein,
  M.~Potthast, Overview of the author identification task at pan-2018:
  Cross-domain authorship attribution and style change detection, in: CLEF,
  2018.

\bibitem{sapkota-etal-2015-character}
U.~Sapkota, S.~Bethard, M.~Montes, T.~Solorio,
  \href{https://www.aclweb.org/anthology/N15-1010}{Not all character n-grams
  are created equal: A study in authorship attribution}, in: Proceedings of the
  2015 Conference of the North {A}merican Chapter of the Association for
  Computational Linguistics: Human Language Technologies, Association for
  Computational Linguistics, Denver, Colorado, 2015, pp. 93--102.
\newblock \href {https://doi.org/10.3115/v1/N15-1010}
  {\path{doi:10.3115/v1/N15-1010}}.
\newline\urlprefix\url{https://www.aclweb.org/anthology/N15-1010}

\bibitem{bagnall}
D.~Bagnall, Author identification using multi-headed recurrent neural networks
  (06 2015).

\bibitem{Qian2017DeepLB}
C.~Qian, T.~He, R.~Zhang, Deep learning based authorship identification, 2017.

\bibitem{inproceedings}
A.~Mohsen, N.~El-Makky, N.~Ghanem, Author identification using deep learning,
  2016, pp. 898--903.
\newblock \href {https://doi.org/10.1109/ICMLA.2016.0161}
  {\path{doi:10.1109/ICMLA.2016.0161}}.

\bibitem{Nirkhi}
S.~Nirkhi, R.~Dharaskar, V.~M. Thakare, Authorship identification using
  generalized features and analysis of computational method, Transactions on
  Machine Learning and Artificial Intelligence (04 2015).
\newblock \href {https://doi.org/10.14738/tmlai.32.1064}
  {\path{doi:10.14738/tmlai.32.1064}}.

\bibitem{azarbonyad2015_time_aware}
H.~Azarbonyad, M.~Dehghani, M.~Marx, J.~Kamps,
  \href{https://doi.org/10.1145/2766462.2767799}{Time-aware authorship
  attribution for short text streams}, in: Proceedings of the 38th
  International ACM SIGIR Conference on Research and Development in Information
  Retrieval, SIGIR '15, Association for Computing Machinery, New York, NY, USA,
  2015, p. 727–730.
\newblock \href {https://doi.org/10.1145/2766462.2767799}
  {\path{doi:10.1145/2766462.2767799}}.
\newline\urlprefix\url{https://doi.org/10.1145/2766462.2767799}

\bibitem{Rexha2018}
A.~Rexha, M.~Kroll, H.~Ziak, R.~Kern, Authorship identification of documents
  with high content similarity, Scientometrics 115 (2018) 223--237.
\newblock \href {https://doi.org/10.1007/s11192-018-2661-6}
  {\path{doi:10.1007/s11192-018-2661-6}}.

\bibitem{Barlas}
G.~Barlas, E.~Stamatatos, Cross-domain authorship attribution using pre-trained
  language models, in: I.~Maglogiannis, L.~Iliadis, E.~Pimenidis (Eds.),
  Artificial Intelligence Applications and Innovations, Springer International
  Publishing, Cham, 2020, pp. 255--266.

\bibitem{shimizu}
T.~Shimizu, Author identification of japanese works using doc2vec and bert-
  proceedings of the 34th annual conference of the japanese society for
  artificial intelligence (06 2020).

\bibitem{Zehfuss}
M.~Zehfuss, War and The Politics of Ethics, 2018.

\bibitem{dower1999embracing}
J.~Dower, W.~N.~. Company,
  \href{https://books.google.co.jp/books?id=gMElGZ93ZXAC}{Embracing Defeat:
  Japan in the Wake of World War II}, W.W. Norton \& Company/New Press, 1999.
\newline\urlprefix\url{https://books.google.co.jp/books?id=gMElGZ93ZXAC}

\bibitem{kyoikushisoshi}
A.~Osada, Kyōiku shisōshi, Iwanami Shoten, 1931.

\bibitem{shinkyoiku}
A.~Osada, Shinkyōiku no kōsō-Amerika no bunka-kyōiku wo hihan shite,
  Fenikkusu Shoin, 1949.

\bibitem{unmei}
A.~Osada, Nihon no unmei to kyōiku, Bokushoten, 1953.

\bibitem{kihonho}
A.~Osada, Kyōiku kihonhō, Shinhyoron, 1957.

\bibitem{NIPS2019_8812_XLNet}
Z.~Yang, Z.~Dai, Y.~Yang, J.~Carbonell, R.~R. Salakhutdinov, Q.~V. Le, Xlnet:
  Generalized autoregressive pretraining for language understanding, in:
  H.~Wallach, H.~Larochelle, A.~Beygelzimer, F.~d'Alch\'{e} Buc, E.~Fox,
  R.~Garnett (Eds.), Advances in Neural Information Processing Systems 32,
  Curran Associates, Inc., 2019, pp. 5753--5763.

\bibitem{zhang2019pegasus}
J.~Zhang, Y.~Zhao, M.~Saleh, P.~J. Liu, Pegasus: Pre-training with extracted
  gap-sentences for abstractive summarization (2019).
\newblock \href {http://arxiv.org/abs/1912.08777} {\path{arXiv:1912.08777}}.

\bibitem{yan2020prophetnet}
Y.~Yan, W.~Qi, Y.~Gong, D.~Liu, N.~Duan, J.~Chen, R.~Zhang, M.~Zhou,
  Prophetnet: Predicting future n-gram for sequence-to-sequence pre-training,
  arXiv preprint arXiv:2001.04063 (2020).

\bibitem{Peters:2018}
M.~E. Peters, M.~Neumann, M.~Iyyer, M.~Gardner, C.~Clark, K.~Lee,
  L.~Zettlemoyer, Deep contextualized word representations, in: Proc. of NAACL,
  2018.

\bibitem{devlin2018bert}
J.~Devlin, M.-W. Chang, K.~Lee, K.~Toutanova, Bert: Pre-training of deep
  bidirectional transformers for language understanding, arXiv preprint
  arXiv:1810.04805 (2018).

\bibitem{lample2019cross}
G.~Lample, A.~Conneau, Cross-lingual language model pretraining, Advances in
  Neural Information Processing Systems (NeurIPS) (2019).

\bibitem{akbik2018coling}
A.~Akbik, D.~Blythe, R.~Vollgraf, Contextual string embeddings for sequence
  labeling, in: {COLING} 2018, 27th International Conference on Computational
  Linguistics, 2018, pp. 1638--1649.

\bibitem{Kestemont2016AuthenticatingTW}
M.~Kestemont, J.~Stover, M.~Koppel, F.~Karsdorp, W.~Daelemans, Authenticating
  the writings of julius caesar, Expert Systems with Applications 63 (2016)
  86--96.

\bibitem{Halvani2019UnaryAB}
O.~Halvani, C.~Winter, L.~Graner, Unary and binary classification approaches
  and their implications for authorship verification, ArXiv abs/1901.00399
  (2019).

\bibitem{PothaStamatatos2018_intrinsic}
N.~Potha, E.~Stamatatos,
  \href{https://doi.org/10.1145/3200947.3201013}{Intrinsic author verification
  using topic modeling}, in: Proceedings of the 10th Hellenic Conference on
  Artificial Intelligence, SETN '18, Association for Computing Machinery, New
  York, NY, USA, 2018, pp. 1--7.
\newblock \href {https://doi.org/10.1145/3200947.3201013}
  {\path{doi:10.1145/3200947.3201013}}.
\newline\urlprefix\url{https://doi.org/10.1145/3200947.3201013}

\end{thebibliography}

\end{document}